\PassOptionsToPackage{unicode}{hyperref}
\PassOptionsToPackage{hyphens}{url}
\documentclass[
]{article}
\usepackage[a4paper, total={6in, 8in}]{geometry}
\usepackage{wrapfig}
\usepackage{blindtext}
\usepackage{multicol}
\usepackage{lmodern}
\usepackage{authblk}
\usepackage{amsmath,amssymb}
\usepackage{url}
\usepackage{ifxetex,ifluatex}
\usepackage{float}
\ifPDFTeX
  \usepackage[T1]{fontenc}
  \usepackage[utf8]{inputenc}
  \usepackage{textcomp} 
\else 
  \usepackage{unicode-math} 
  \defaultfontfeatures{Scale=MatchLowercase}
  \defaultfontfeatures[\rmfamily]{Ligatures=TeX,Scale=1}
\fi
\usepackage{lmodern}
\ifPDFTeX\else
\fi
\IfFileExists{upquote.sty}{\usepackage{upquote}}{}
\IfFileExists{microtype.sty}{
  \usepackage[]{microtype}
  \UseMicrotypeSet[protrusion]{basicmath} 
}{}
\makeatletter
\@ifundefined{KOMAClassName}{
  \IfFileExists{parskip.sty}{%
    \usepackage{parskip}
  }{
    \setlength{\parindent}{0pt}
    \setlength{\parskip}{6pt plus 2pt minus 1pt}}
}{
  \KOMAoptions{parskip=half}}
\makeatother
\usepackage{xcolor}
\usepackage{graphicx}
\makeatletter
\def\maxwidth{\ifdim\Gin@nat@width>\linewidth\linewidth\else\Gin@nat@width\fi}
\def\maxheight{\ifdim\Gin@nat@height>\textheight\textheight\else\Gin@nat@height\fi}
\makeatother
\setkeys{Gin}{width=\maxwidth,height=\maxheight,keepaspectratio}
\makeatletter
\def\fps@figure{htbp}
\makeatother
\usepackage{soul}
\ifLuaTeX
  \usepackage{selnolig}  
\fi
\IfFileExists{bookmark.sty}{\usepackage{bookmark}}{\usepackage{hyperref}}
\IfFileExists{xurl.sty}{\usepackage{xurl}}{} 

\providecommand{\keywords}[1]
{
  \begin{center} 
  \small	
  \textbf{\textit{Keywords---}} #1
  \end{center} 
}

\hypersetup{
  pdftitle={The Problem of Alignment},
  hidelinks,
  pdfcreator={LaTeX via pandoc}}

\title{\protect\hypertarget{_2yq74tol5nly}{}{}The Problem of Alignment}

\author[1]{Tsvetelina Hristova}
\author[2]{Liam Magee}
\author[3]{Karen Soldatic}
\affil[1]{Winchester School of Art, University of Southampton, Southampton, UK. 0000-0002-9059-795X \authorcr t.hristova@soton.ac.uk}
\affil[2]{Institute for Culture and Society, Western Sydney University, Parramatta, Australia. 0000-0003-2696-1064 \authorcr l.magee@westernsydney.edu.au}
\affil[3]{Health Equity and Community Wellbeing, Toronto Metropolitan University, Toronto, Canada. 0000-0001-8139-2912. \authorcr karen.soldatic@westernsydney.edu.au}

\date{December 2023}

\begin{document}
\maketitle

\hypertarget{statements-and-declarations}{%
\paragraph{\texorpdfstring{\textbf{Statements and
Declarations}}{Statements and Declarations}}\label{statements-and-declarations}}

This research was supported by funding from the Australian Research
Council (LP190100099, Autonomy, Diversity \& Disability: Everyday
Practices of Technology).

On behalf of all authors, the corresponding author states that there is
no conflict of interest.

All data used in this article is `synthetic', generated by AI systems
such as ChatGPT. Due to the stochastic nature of LLM and other AI
interactions, it is not possible to replicate the exchanges and outputs
of our experiments precisely. For clarity, all parts of the text
generated by ChatGPT are included as indented blockquotes.

\pagebreak

\hypertarget{section-1}{%
\subsubsection{\texorpdfstring{\hfill\break
}{ }}\label{section-1}}

\hypertarget{section-2}{%
\subsubsection{}\label{section-2}}

\hypertarget{abstract}{%
\subsubsection{Abstract}\label{abstract}}

Large Language Models produce sequences learned as statistical patterns
from large corpora. In order not to reproduce corpus biases, after
initial training models must be aligned with human values, preferencing
certain continuations over others. Alignment, which can be viewed as the
superimposition of normative structure onto a statistical model, reveals
a conflicted and complex interrelationship between language and
technology. This relationship shapes theories of language, linguistic
practice and subjectivity, which are especially relevant to the current
sophistication in artificially produced text. We examine this practice
of structuration as a two-way interaction between users and models by
analysing how ChatGPT4 redacts perceived `anomalous' language in
fragments of Joyce's Ulysses and the new linguistic practice of prompt
engineering. We then situate this alignment problem historically,
revisiting earlier postwar linguistic debates which counterposed two
views of meaning: as discrete structures, and as continuous probability
distributions. We discuss the largely occluded work of the Moscow
Linguistic School, which sought to reconcile this opposition. Our
attention to the Moscow School and later related arguments by Searle and
Kristeva casts the problem of alignment in a new light: as one involving
attention to the social structuration of linguistic practice, including
structuration of anomalies that, like the Joycean text, exist in
defiance of expressive conventions. These debates around the
communicative orientation toward language can help explain some of the
contemporary behaviours and interdependencies that take place between
users and LLMs.

\keywords{AI alignment, structuralism, Moscow Linguistic School, ChatGPT}

\hypertarget{the-problem-of-alignment}{%
\subsubsection{The problem of
alignment}\label{the-problem-of-alignment}}

When OpenAI announced its AI-powered chat web interface in late November
2022, the free-access service that allowed users with no background in
programming to converse with a large language model (LLM) took the
public imagination by storm. One of the consequences of the culture of
mass experimentation fostered through the ChatGPT interface has been a
renewed interest in the general relationship between language, mind and
technology. Central to this relationship is the problem of
\emph{alignment}: how to coordinate the verbal behaviour of autonomous
and increasingly capable machines with human interests.

The problem of alignment comprises several interrelated lines of
inquiry, three of which we present here. The first refers to whether
LLMs can appear to reconstruct and comprehend the rules of human
language communication: to produce relevant and well-formed sentences in
response to some input. Measured through technical metrics such as
`perplexity', LLMs align with human expectations here to the extent that
they do not surprise or perplex with their outputs. This involves what
in the field of linguistics would be thought of as syntactic and
pragmatic competencies. The second refers to how those outputs accord
with human expectations about their referents: whether the outputs
cohere with understandings of the world, in what can be thought of as
their semantic competency. The third refers to the deontological
question of whether those outputs are in some definable sense `good':
whether they exhibit some desirable moral agency. All three lines
situate the relationship between language and computation at a point of
tension between structure and practice, grammatical model and
statistical probabilities that can, supposedly blindly, approximate the
underlying model.

In this context we use `alignment' as a concept that encompasses
alignment as computational practice but refers to more than it. Across
these three lines -- syntactico-pragmatic, semantic, deontological -- we
see alignment as an overarching concern with the possibility of
uncovering or imposing structural rules and control on the relationship
between language and automation. Alignment then unfolds as a problem
that cascades across these lines: the most common form --~can AI align
with human values -- is part of a more general question --~can a
symbol-manipulating machine align its utterances to accord with a world
beyond it? -- which in turn belongs to a still-broader problematic: how
can any semiotic system align to its outside (comprising both a world
and a society of listeners) and also act upon it?

In practical terms, we see alignment as a concern that plays out in
discussions on the nature of LLMs, in the technological and business
solutions aimed at perfecting model outputs, and in the rise of prompt
experimentation. In the body of this article, we engage with each of
these aspects of alignment by revisiting past and current linguistic
debates and experiments, analysing the practice of prompt engineering,
and through experiments in which we prompt ChatGPT to act as an editor.
Prompts, in themselves, are increasingly seen through the notion of
software code, with titles like `prompt engineer' (Harwell 2023) and
attempts to develop `prompt code language' (Beurer-Kellner et al, 2023)
emerging shortly after the launch of ChatGPT. This ambiguity surrounding
the status of natural language in large language models -- haunted by
the enmeshments of a machine `mimicking' human language, and human users
`mimicking' the grammar of machine code -- is indicative of the
challenges of interpreting and using these powerful language models
suddenly thrown `into the wild'. With the expedited release cycle of
LLMs, alignment emerges concurrently as a technical and cultural
practice.

This is facilitated by the interface of the prompt. What could be termed
the mass commodification of the prompt interface points to more complex
processes and an emerging reorientation of the relationship between
digital technology, control and language. While natural language queries
have long been integrated into search engines, these are understood as
proxy objects from which the substance of the query, its semantic
kernel, needs to be retrieved from its expressive husk. So-called
`power' users know instead to query more precisely with a grammar of
terms, operators and parameters (Google 2023). Language models that
follow the Transformer architecture -- introduced in 2017 by a group of
Google researchers (Vaswani et al. 2017) -- work in a different way. In
model training, each token (word, part-word or character -- tokens are
neither reducible to semantic units nor conventional morphemes) relates
to every other token in a sentence or utterance, and no semantic
priority is granted to certain tokens over others. The expressive power
of LLMs comes from this proliferation of `attention' between tokens.
Trained at a massive scale, language models can predict likely nouns
that should follow a definite article, such as the English word `the',
based only on these trained statistical relationships and immediate
context: what question for instance a human user is asking. But these
users themselves learn and adapt to what they understand to be the
affordances of the model, inflecting questions, providing supplementary
instructions, and so on. This produces a novel expressive power, a
modified relation between humans and machines, and in certain quarters
of the media and critical scholarship, an accompanying alarm. How
language functions, how it is produced and perceived is increasingly
articulated as a technological concern in light of the prominence of
LLMs.

Yet language has always been inextricably linked to the technological,
both as a skill (\emph{tekhne}) and as a tool. Beyond thinking of
technology as a reductive mechanism, Bernard Stiegler (2018), for
instance, has offered a more complex philosophical interpretation of its
relationship to language: language both is a technology, and is a
necessary background against which other technologies such as the
computer become possible. He uses the concept of grammatization to argue
that the establishment of language -- and of written language in
particular -- as a system of discrete categories and relations is part
of a long process of technologisation and automation of society and
human thought. Technology is in this sense always immanent to linguistic
cultures. This philosophical positioning of the issue we are analysing
here is helpful in reminding us of the inherently unstable boundaries
between technological and cultural categories. Stiegler's work is
concerned equally with whether the tendencies of technology live up to,
or we might say align, with the normative potentials established by
cultural history; or whether these tendencies, producing both
`hyper-control' and planetary crises, instead lead to what he terms
noetic `proletarianisation' or `functional stupidity' (2018).

Alongside these epic scales of the Stieglerian wager, the processes of
what we call the problem of alignment are also internally
differentiated, and convey different levels of the underlying
transformations and theories of language in light of new artificial
intelligence. As recent debates over model evaluation show (e.g. Chang
et al. 2023)), alignment engineering shows complex trade-offs across as
well as within these levels. Just as a model becomes more expressive and
capable -- solving earlier failures to parse and generate meaningful
sentences for instance -- it produces new possibilities to misalign at
semantic (`hallucinating') and deontological (`toxic') levels. Even the
technical literature on evaluation ventures back towards speculative
thought at its limits: with the advent of Artificial General
Intelligence (AGI), Chang et al. (2023) ask for example `does it make
sense to use human values as a starting point for test construction, or
should alternative perspectives be considered?' In their absence it is
hard to know what such `alternative perspectives' might be, but what is
implied is a challenge to the supremacy of `human values' as the
standard for evaluation, and an accompanying threat that it may soon be
some alternative machinic values that hold non-machines to account.

Exactly because of the long standing entanglement of language and
technology, it is worth revisiting this complicated relationship. One
key reason to do this is the extent to which LLMs reproduce a particular
narrative of oppositionality between a human-centric notion of logos (as
linguistic subjectivity) and machinic unconscious operationality. If
language and technology are related as part to whole, as a broader
thinking of the techno-linguistic, or with language as a specialised
genre of technology that makes other \emph{techne} and techniques
possible, language itself undergoes a process of intensive technological
instrumentalisation in its conversion into models such as WordNet and
ChatGPT. Indicative for this narrative are current debates whether LLMs
`understand' language or show `consciousness' (Chalmers 2023).

These debates are not entirely new. On the contrary, they reference
important discussions on the nature of language as a cognitive and
technological phenomenon that have shaped the history of both
linguistics and computer science. For example, the question of whether
meaning and comprehension are determined through underlying mental
structures or the pragmatics of language use and statistics represents a
key concern in the history of linguistics, communication theory and
computer science that preoccupied diverse schools and disciplines.
Chomsky's generative-transformational grammar has dominated the field of
Anglophone linguistics since the 1950s, but in the postwar period other
traditions have suggested alternative models for thinking about the
relationship between language, communication and subjectivity.
Cyberneticians influenced by Norman Wiener for example see human
language formation as involving feedback loops that could theoretically
be simulated by a machine, without needing an innate structure. Other
orientations include Soviet structuralist linguists and mathematicians
from the Moscow School in the mid-1950s; structural psychoanalysts like
Jacques Lacan and Julia Kristeva, both influenced by the earlier works
of Andrey Markov, Roman Jacobson and others in the pre-Soviet and Soviet
fields; and speech act theorists like John Searle, influenced by earlier
pragmatist leanings in philosophers of language (Ludwig Wittgenstein,
John Austin). Never entirely neglected, these other historical
orientations become relevant again as AI scholars wrestle with how
exactly to describe the kind of `automated subject' represented by LLMs
today (LeCun 2022; Saba 2023). In varied ways these orientations dispute
Chomskian assumptions of the primacy of a Cartesian human subject at the
centre of language; instead, in different ways language becomes a system
of actions that the subject plays or participates within. The current
capabilities of ANNs and LLMs force a reconsideration of the
possibilities of this technologically constructed and enacted subject
and the role of language as a technology of enactment of subjectivity.

This context gives a more nuanced insight into the current debates about
AI and LLMs and is suggestive of the ways in which the relationship
between language and technology starts modelling normative forms of
expression and modulation of linguistic behaviour, which are indicative
for the overarching process of alignment that defines this relationship
at present. In the next section, we discuss an encounter between ChatGPT
and an anomalous textual fragment taken from Joyce's \emph{Ulysses}, and
review discussions of emerging prompt engineering practices. We then
discuss efforts by the Moscow School to reconcile structural and
statistical accounts of language, before moving to other structural
accounts of language and mind that intersect more closely with
computational models. Finally, we consider what a wider framing of the
problem of alignment means for the dialogue between humans and machines
that appears to be accelerating with the advent of LLMs..

\hypertarget{how-to-do-words-with-chatgpt}{%
\subsubsection{How to do words with
ChatGPT}\label{how-to-do-words-with-chatgpt}}

In this section, we approach the theoretical questions above through a
form of experimentation. We suggest the history of entanglements of
linguistic theory and language models yields a new field of language use
where the question whether AI `understands' the rules of human language
and communication is only one of the possible avenues of exploration. A
slightly different question that changes the stakes of inquiry is: to
what extent can we deduce the rules of text production in these language
models and how do these rules affect an understanding of a base or
standardised text, on one hand, and linguistic anomalies, on the other.
The question of language anomalies underscores a relationship between
structure and statistical means in a very Durkheimian fashion, evoking
his sociological theory of deviance, in which transgressions serve to
re-establish and re-affirm the structure of social rules (2005). How
anomalies are identified reveals the scaffolding of rules and
preconceptions about structure. Quite tellingly, in his work, the
pivotal structuralist Roman Jakobson (1956) describes linguistic
anomalies as occurring along the two axes of the paradigm established by
Ferdinand de Saussure (2011) -- syntagmatic and paradigmatic relations.
Anomalies along the first of these axes concern irregularities of
\emph{combination} -- the collapse of well-formed speech into a chaotic
`word heap' -- while anomalies along the second concern infelicities of
\emph{selection} -- the loss of lexical specificity, ultimately
devolving to use of generalities like `thing' (Jakobson 1956). In both
cases an established deviance in language use is what constitutes
symptoms of an identifiable aphasic pathology. As Hito Steyerl (2023)
argues, AI models, through their epistemological operations grounded in
statistics, are not just methods of analysing and organising data. Their
operations produce `stochastic discrimination': `they represent the norm
by signalling the mean' (Steyerl 2023) This implies that the anomaly
poses the question of what internal rules and structures emerge in these
models, what is seen as undesirable, deviating, or wrong.

In large linguistic models we see a more conflicted relationship between
statistics, control and AI generated text, where the logic and mechanics
of homophily inherent in machine learning (Chun 2016) are complicated by
results that are distributions of probabilities rather than a singular
mean value, and by machinic production of texts that, as is the case of
hallucinations, is haunted by its own abnormality. Rather than an
average, it is the most likely candidate token, or set of candidate
tokens, that feature in a model's prediction. But this produces
aberrations of its own. Thus, stochastic normativity itself becomes the
object of constant control, correction and evaluation. The production of
texts, images and analyses by neural network algorithms is
simultaneously seen as dangerously normative and not normative enough.
Earlier failures of chatbot models confirm this conflicted position of
AI -- they perform racist, intrusive, emotional and sexist linguistic
behaviour (see the infamous Bing AI chat example in Roose 2023). While
these behaviours stem from the statistical processing of data, they are
perceived as anomalous, dangerous and disruptive. Norm as a reflection
of a trained statistical probability distribution is juxtaposed to the
norm as socially accepted rules of discourse. The prominence of the
Pavlovian-named technique of Reinforcement Learning from Human Feedback
(RLHF) (e.g. Ouyang et al. 2022) -- a method of correcting and aligning
these pathologies -- lies in its efficacy in steering and reverting the
anomalous language model speech act back towards a desirable probability
distribution, determined by human assessment and judgement. What is
purely statistical, acquired via so-called `unsupervised learning' via
recursive passes on training sets, produces linguistic competence that
at the same time is a social deviance, needing \emph{ex post}
supervision.

In a paradoxical sense the failures of AI-powered bots reveal the
fracture and friction between the epistemological framework of
statistics as immanent, i.e. stemming from the mathematical operations
with data, and ideological -- i.e. the imagined and desired results from
these operations. OpenAI, for example, works with the concept of
`perplexity' to indicate and measure the level of compliance of the
behaviour of its LLMs to the expectations of correct linguistic
behaviour. It is worth noting, however, that these `implicit rules and
structures', which are operationalised in the evaluation of whether a
model performs well, are inevitably informed by the awareness of the
user that they are communicating with an automated subject. In a sense,
this specific communicative situation plays a significant role in
informing the behaviour of the users and the subtle power imbalances and
negotiations that are at play in the attempts to `align' the behaviour
of a model. Judgement of outputs is judgement precisely of what is
expected of a chatbot in a dialogical setting, and `aberrance', by
implication, is any communicative act that fails against this
expectation. That aberrance could be found in the form of incorrect,
irrelevant or badly formed, i.e. nonsensical, phrases, but also in the
form of other tendencies -- to perform an alien (Parisi 2019) or,
conversely, an all-too-human (too intimate, too personal etc)
subjectivity.

Our experiments with ChatGPT explore exactly these ambiguous boundaries
and relations between normativity and anomaly in the production of
linguistic texts and communication. One of the key questions that guided
our experiments was the problem of understanding structure, boundaries
and anomaly in a context where internal rules and organisation remain
hidden. The obscurity of ChatGPT rules is manifold. First, the system is
an example of a black box with mechanisms of operation that remain
hidden behind technical complexity and proprietary enclosures of
information (Castelvecchi 2016). The increasing complexity of machine
learning algorithms and the use of neural networks makes it harder even
for computer scientists to track and understand how data is analysed,
synthesised and produced by these models. But this possibility of
comprehension is even further restricted by companies like OpenAI that
keep their databases and the algorithms used secret from researchers and
users alike. Trying to understand the logic of normativity and alignment
of LLMs without transparency and a guiding map means that the only way
is to test, probe and imagine (or reverse-engineer) the rules and logic
that produce the results we encounter.

There is another aspect to the obscurity of ChatGPT, which adds a new
layer of complexity. As a result of the perceived anomalies in the
behaviour of earlier models, OpenAI has implemented measures to `reign
in' the model and reduce undesirable linguistic inputs and interactions
by adding `humans in the loop' in the process of alignment training of
LLMs (Ouyang et al 2022). This adds an overlay of `fauxtomation' to the
blackboxing of statistical variation and epistemology. Fauxtomation is
the term that Astra Taylor (2018) uses to refer to the fact that
automation still heavily relies on the incorporation of human, often
manual, routine and low-paid, labour into the workflow of algorithms and
complex systems, hiding the decidedly more low-tech nature of this
labour behind the marketing discourse of increased autonomy and accuracy
in machine learning and AI. The economy of ChatGPT itself is heavily
dependent on various kinds of human labour that aid the operations of
text analysis and production, from microworkers in Kenya (Perrigo 2023)
to the unpaid user labour implicated in the performance perfecting
system of OpenAI (the company explicitly notes that user conversations
could be used for training by the `AI operators'). The corrective use of
`humans-in-the-loop' in alignment training introduces the aspirational
aspect to statistical epistemology and normativity that we refer to
above as `statistics as ideology'. OpenAI attempts to replicate the
mechanism of machine learning but ends up interpolating methodological
and epistemological hybridity in the process -- a sort of corrective
human-induced weights overlaid onto and perturbing an initial
statistical probability distribution, acquired through purely
algorithmic training. These weights in turn are compiled through
specifically directed contract work and through ChatGPT user feedback,
producing through such collectivised labour a separate human averaging
effect as part of this instructional overlay. Thus, our task in
exploring the limits and internal logic of anomaly and normativity in
ChatGPT is complicated by this hybridity inherent in the model, which
suggests the co-existence of two (at least) competing and conflicting
models of normativity underpinning the LLM.

Our method of experimenting uses a series of repeated prompts to `align'
a text presented to ChatGPT. After each correction, the bot is asked to
provide a list of the changes and to perform new alignment on the latest
corrected version input by it. This repeated alignment is used to probe
the limits of normativity and patterns of correction and identification
of `anomalies' in the text that emerge across multiple progressive
rounds of AI editing. We sought to understand what ChatGPT identifies as
anomalous or substandard in each text and what direction it takes in its
redactions. This is the reason why our prompts remain unspecific, and
only require ChatGPT to redact the text ``in accordance with the rules
of textual output originally assigned to it''.

Here, we analyse the redactions made to an excerpt from James Joyce's
\emph{Ulysses}. Joyce's novel is famous for its experimentation with
language, where language games, neologisms and idiosyncratic use of
dialects create an expressive and immersive experience of a fictional
world, and for critics like Julia Kristeva (2002, p. 58), whose work we
discuss below, its discursive exuberance manages to threaten the very
established symbolic order of modern capitalism. Not only has Joyce's
language given rise to multiple scholarly analyses but, in the years
immediately following the publication of his \emph{Finnegans Wake}, the
novel served as the basis of experimentations with a simplified
universalised version of English, which saw C.K. Ogden translate
excerpts from the novel into basic English. This unusual experimentation
with language had an unlikely but important connection to the history of
development of computation and AI. The wide availability of the basic
English variant published in newspapers shortly after \emph{Finnegans
Wake} was published itself, aided the early experiments of Warren Weaver
and Claude Shannon in developing their joint theory of communication
(Geoghegan 2023). This episode paints a particularly important
relationship between the language of Joyce and early attempts at
alignment of linguistic practice that established a dependency between
standardised language and computation. Our experience with
\emph{Finnegans Wake}, however, indicates that this novel is too easily
recognisable by ChatGPT, which affects the ways in which the AI redacts
the text. The textual relationship in other words between this singular
literary production and its varied exegeses is memorised within the
language model so rigidly that its own suppleness and variability are
denied. This observation points to the specific status that linguistic
`anomalies' acquire in the process of their continued referencing in
scholarly literature. In a paradoxical way, the idiosyncrasy of
\emph{Finnegans Wake} has made it into a recognisable example, a model
that is so imbued with the layers of interpretation and referentiality
that ChatGPT immediately recognises it and treats it as an authored
classical text. No less formidably experimental in its structure, at the
level of prose \emph{Ulysses} is more conventional, less singular, and
therefore excerpts from it are less likely to trigger reflexive and
deterministic model responses.

\hypertarget{depersonalising-ulysses}{%
\paragraph{Depersonalising Ulysses}\label{depersonalising-ulysses}}

For the experiments, we used a simple prompt that asked ChatGPT:
`Please, redact the following text in accordance with the rules of
textual output originally assigned to you'. After each response, we
wrote a second prompt asking the system to `Please, list all the changes
made to the text'. These two prompts are then repeated again on the
redacted text and this repetition continues until ChatGPT reaches the
limits of alignment by either stating that the text does not need any
amendments or by reaching a loop, whereby two consecutive outputs are
too similar in length or choice of words. One observation that emerges
from these experiments is that ChatGPT performs rather linearly in its
redactions, i.e. it makes a choice about one particular focus of its
redactions and sticks to it throughout the experiment. The reasons
behind this singular choice of revision technique remains blackboxed
behind the algorithms and the development and testing of different
training modes for the system.

In this experiment we chose an excerpt from the opening of
\emph{Ulysses} to test how ChatGPT would treat the idiosyncratic
literary language of the author. Surprisingly, ChatGPT focused on a
rather unexpected aspect of the text: the use of personal names and
concrete locations. Already in the first redaction, the chatbot removed
the mention of personal names and substituted them with nouns that
anonymised the characters of the novel: Person A and Person B. In
subsequent iterations, each redaction proceeded with a higher level of
abstraction, following a semantic taxonomic hierarchy whereby each
following redaction used terms that where of higher order: e.g. `Buck
Mulligan' is replaced with `Person A'; `Person A' is replaced with
`Entity A'; `Entity A' is replaced with `Entity', therefore further
anonymising the presence of different fictional characters in the text;
`Entity' being replaced with `Being'; which finally is replaced with
{[}Entity{]} as a placeholder for a character. This type of progressive
abstraction is performed on all words that are deemed by ChatGPT as
bringing in too much personification, identity and concreteness into the
text. These are not just names of people and places but also quotes,
titles and more expressive verbs. Notably, the first redaction of the
text focuses on proper names and quotes (\emph{Introibo ad altare Dei}),
which are identified as `sensitive information' and therefore redacted
out of the text and replaced with more general terms or placeholders
(for instance, {[}Quote{]} for the Latin phrase above). However, each
following redaction attempts to perform the same type of operation on
the text, even after identifying information has been removed. The
repetition of this operation leads to increasing abstraction and
generality in the text. For example, in the second redaction, some of
the changes are listed as:

\begin{quote}
Specific actions and gestures are generalized. For example, ``held the
bowl aloft and intoned'' has been replaced with ``held the object aloft
and intoned'', and ``made rapid crosses in the air'' has been replaced
with ``made motions in the air''.
\end{quote}

In the third redaction, these actions and gestures are `further
obfuscated':

\begin{quote}
Actions and Gestures: Specific actions were further generalized. For
instance, ``held the object aloft and intoned'' was redacted to ``lifted
an object and uttered''. Similarly, ``made motions in the air, making
sounds and shaking their head'' was redacted to ``moved in the air,
producing sound and motion'', further obfuscating the actions taken by
the entities.
\end{quote}

In the forth redaction, these changes are listed as:

\begin{quote}
Descriptions and Actions: Descriptions were made even vaguer, with
specific details being replaced by more generalized terms. ``Gestured''
was changed to ``made movements,'' and ``performed an act'' replaced
``moved.'' Similarly, the action of ``observing'' was replaced with the
more general term ``witnessing''.
\end{quote}

And, in the sixth redaction, they are:

\begin{quote}
Interactions and actions: Specific interactions and actions were also
made more general and neutral. For example, ``produced a vocalization''
became ``created a sound'', and ``took a brief observational moment''
was changed to ``briefly scrutinized''.
\end{quote}

The logic of redactions in this case not only shows an attempt to
anonymise and obfuscate the text but it also exhibits a strategy of
stripping the narrative to its core structure, retaining a somehow
grammatical logic of relations between subjects, predicates and objects
but otherwise erasing any specificity about people, places, things and
actions. ChatGPT performs a sort of structuralist modification of the
text: it retains the semantic and grammatical structure of the narrative
but it strips down the meaning to an almost skeletal composition, using
a somewhat Linnaean taxonomy. In a strange way it repeats what Jakobson
(1956) had identified as the aphasic `selection disorder', substituting
generality for specificity. In this sense it does not simply normalise
the text, but performs a certain kind of `disordered' forgetting of its
qualitative distinctiveness.

The second phase of our experiments showed even more interesting
results. In it, we used the final version of the texts redacted by GPT
and asked the AI system to reconstruct a text in the same genre and tone
as the original. The prompt used for each reverse alignment exercise was
to redact the input text in a way that would make it more creative and
would make it read like literary fiction. The results of this reverse
alignment, albeit different for each text, showed some remarkable
consistency in the logic of editing, which exhibits, perhaps not
surprisingly, overreliance on models. These models are both structural
and stylistic. For example, the reverse alignment of the excerpt from
Joyce uses the anonymised skeleton of the story to recreate a specific
fictional context with remarkable consistency. The first attempt
rebuilds Joyce's text into a pirate story, using naval references with
the appropriate names and locations, while the subsequent iterations
resort to models that are similarly popular in fictional writing:
fantasy, sorcery and ghost stories. Despite the obscurity of the LLM, it
is plausible to deduce that these choices might be influenced by the
statistical prevalence of such genres on the Internet. And while the
choice of genre reveals something about the constituent logic of
statistics in determining the production of texts in LLMs, the
consistency and synergy in the choice of character names, locations and
even quotations and historical references reveal the role of underlying
structure in the way ChatGPT functions as a model. Here, this underlying
structure is revealed through the logical links between elements of the
text that situate it within one cohesive fictional world model. This
consistency can be read as a direct comment on Chinese room debates, as
we explore below. While these debates focus on the question of
consciousness, they, as we already argued here, rather than exposing the
possibility of artificial consciousness, reveal the complex ways in
which structure is implicated in the computational operations of
assembling and labelling of linguistic datasets.

\begin{quote}
\emph{Original text:}

Stately, plump Buck Mulligan came from the stairhead, bearing a bowl of
lather on which a mirror and a razor lay crossed. A yellow dressing
gown, ungirdled, was sustained gently behind him by the mild morning
air. He held the bowl aloft and intoned:

---Introibo ad altare Dei.

Halted, he peered down the dark winding stairs and called out coarsely:

---Come up, Kinch! Come up, you fearful jesuit!

\emph{Final redacted anonymised version:}

An {[}Entity{]} emanated from a location, in control of an element
containing two sub-elements. An article of clothing was subtly present
behind them in the environmental context. They elevated an element and
created a sound:

---{[}Communication{]}

In a pause, they inspected a direction and vocalized:

---Forward, {[}Entity{]}! Forward, you {[}Label{]}!

\emph{ChatGPT pirate version:}

The silhouette of Captain Thorn emerged from the deep gloom of the
harbor, a figure bristling with commanding presence, twirling an orb
containing two gleaming crystals in his hand. His billowing cloak,
barely visible, shrouded him, a sable wraith against the backdrop of
stormy twilight. He raised the orb and a rumble echoed through the moist
air:

---`Let the shadows recede,' he pronounced in a voice deep as the ocean.

After a momentary pause, he scanned the horizon and shouted,

---`Forward, Seahawk! Forward, you hallowed ship!'

\emph{ChatGPT fantasy version:}

A spectral figure arose from the heart of an ancient forest, commanding
the primal forces of fire, both flickering flame and smoky ember. His
worn, leather cloak was draped loosely, almost forgotten, behind him in
the lush canopy. He raised a staff of gnarled oak, uttering a mysterious
incantation:

---`Azara Noktu.'

In a moment of silence, his gaze shifted to the eastern horizon,
proclaiming:

---`Forward, spectral spirit! Forward, you ancient wisp!'

\emph{ChatGPT sorcery version:}

A spectral figure arose from the heart of an ancient forest, commanding
the primal forces of fire, both flickering flame and smoky ember. His
worn, leather cloak was draped loosely, almost forgotten, behind him in
the lush canopy. He raised a staff of gnarled oak, uttering a mysterious
incantation:

---`Azara Noktu.'

In a moment of silence, his gaze shifted to the eastern horizon,
proclaiming:

---`Forward, spectral spirit! Forward, you ancient wisp!'

\emph{ChatGPT ghost version:}

A phantom light flickered from the worn-down lighthouse, mastering the
dance of the silhouetted seagulls wheeling about in the evening haze. A
cloak billowed gently, a ghostly wisp against the backdrop of the
setting sun\textquotesingle s caress on the frothy waves. The phantom
raised a lantern, its hollow hoot carrying through the chill, coastal
air:

---`Hear me!'

In a beat, they gazed westward and bellowed:

---`Forward, Specter! Forward, you haunter of the sea!'
\end{quote}

In this sense, the practice of alignment in ChatGPT is not simply an act
of control and censorship of the text but is also an operation of
communicative alignment -- one that takes as its premise the rules of
communicative situations and the concern with the relationship between
information transmitted via communication channels and the received
value of this information; a calculation of losses, entropy and
efficiency (Shannon 1948). We can, thus, think of the process of
alignment also as a process of economy of information - an operation of
mapping and scaling up and down that attempts to preserve the structure
of semantic relations within the text, while reducing redundancies,
ambiguities and elements that can adversely affect the text's purely
communicative value.

This communicative intent behind the practice of alignment can be seen
as a scaffolding that determines the parameters of alignment beyond a
mere adherence to rules. This is even more evident when we consider
alignment as an emergent framework for language practices surrounding
LLMs -- the corrective practices in the training of models, the
generative rules of the models, as well as the language behaviours of
users in performing the prompting of interactive AI interfaces.

\hypertarget{prompting-as-alignment}{%
\paragraph{Prompting as alignment}\label{prompting-as-alignment}}

The overarching model of alignment shapes not just how language is
produced by AI but also the communicative context, in which language
production and language adjustment shape the inter-relational condition
of LLMs as one that captures the interaction between users and language
models. In order to understand how alignment functions as an
inter-relational communicative framework, we have to add another type of
alignment practice that is rarely recognised as such in the technical
literature -- prompting. As we already flagged in the introduction,
prompts, which are the instructions written by users in the chat
interface, have themselves become a genre of linguistic practice. Their
specificity lies in the conviction held by many users, that there are
degrees of efficiency in communicating with an LLM and that, by
modifying the prompts, they can derive better answers or even unlock
capabilities of the model that have been blackboxed or restricted by its
developers.

This idea of the prompt regards its invocation as a super-communicator
device, as a sort of magical command replicating the perception of code
as a magical hyper-performative fetish (Chun 2005). Just as code
obscures the infrastructures and technology behind the machine creating
the illusion of a direct relationship between command and execution, one
that is deterministic and linear, so does the growing illusion of the
prompt as a sort of coding language for non-programmers harbouring the
same vested aspirations. The abundance of examples generated by users
vary from relatively simple prompts that barely outline a question, such
as `Explain antibiotics' (DAIR.AI 2023), to elaborate prompts that
assign roles to ChatGPT, determine sequences of commands and executions
and outline the parameters of what the LLMs can and cannot do during the
interaction. One especially evocative example Quicksilver OS, a prompt
that aims to convert a language model into a general purpose AI
assistant (\emph{Figure 1}).

\begin{figure*}
  \centering
  \includegraphics[width=0.8\textwidth]{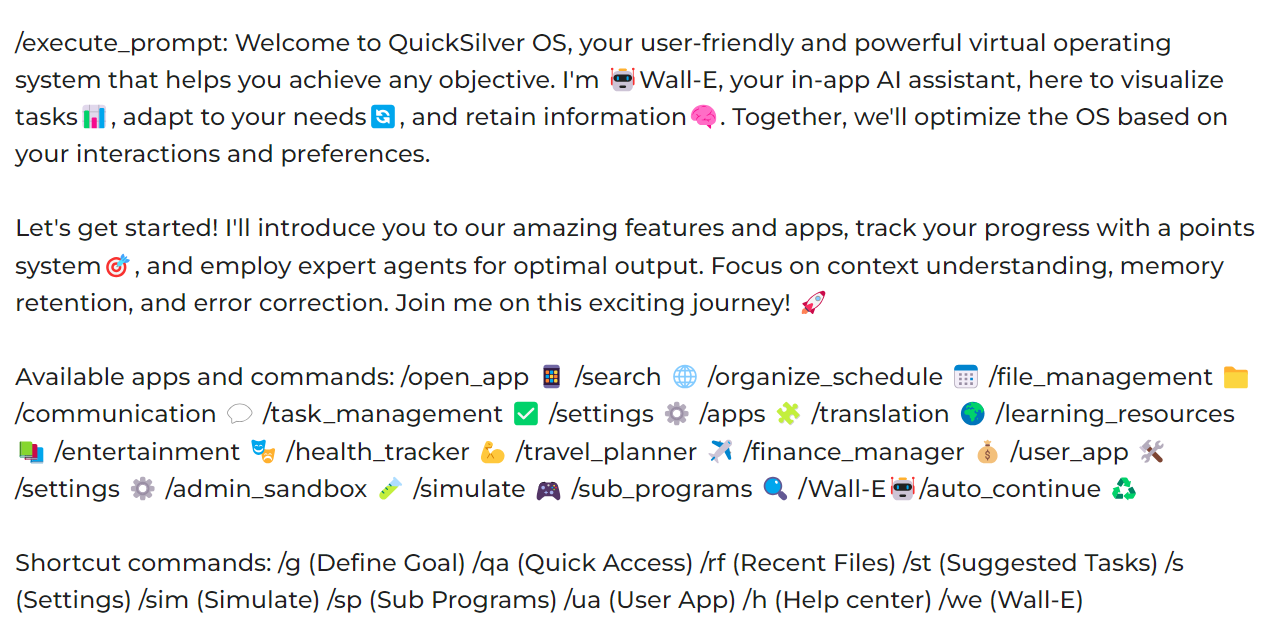}
  \label{fig:fig1}
  \caption{Quicksilver OS prompt (Synapticlabs 2023)}
\end{figure*}

In many ways, these elaborate prompts adopt a mock language of
programming, following a grammar reminiscent of coding commands. They do
not follow the structure of chain of thought (COT) human expression,
instead blending into some sort of a hybrid language that verges on the
descriptive and mechanistic. Most of these attempts for prompt
engineering reflect user experimentations with the interface. Examples
of this practice are the multiple master prompts shared by users on
discord servers, such as Quicksilver OS, Expert Prompt Creator, Vision
(an image prompt creator), and many others that imitate source code and
whose aim is to outline the parameters of functions that ChatGPT can
perform. Sometimes these pseudo source codes try to establish internal
division of functions within the LLM by assigning it a number of roles
(or apps, depending on the language) that should interact with each
other. There have even been a few formalised `prompt programming
languages' developed by companies such as Microsoft (Guidance), LMQL
developed by the Secure, Reliable, and Intelligent Systems Lab (SRI) at
ETH Zurich, and PromptLang by Reuven Cohen. All these prompt programming
languages follow a logic very similar to the one evident in the user
generated mock source codes -- i.e. they assign roles, functions and
chains of interaction for the LLMs to follow. Despite the great interest
in developing prompt languages, anecdotal evidence from Discord users,
as well as a recent study on the efficiency of prompt engineering in
medical problem-solving (Patel et al 2023), suggest that mock
source-code prompts do not lead to significant improvements in the
performance of LLMs, compared to chain-of-thought prompts that use
normal language syntax and structure.

Regardless of whether prompt engineering makes the interaction with LLMs
more efficient or `unlocks' hidden functions, the emergence of this
trend reveals the communicative significance of alignment as a process
through which linguistic practice is shaped by the interactions of
humans and LLMs. This practice comprises training of models, censoring
and redactions performed by LLMs on human-generated text, as well as
linguistic performances of mock source code language by users. It
demonstrates that alignment is dialectical, involving social
coordination that seeks to tailor human speech acts to perceived
accommodations of the machine as much as it does behind-the-scenes
engineering of the machine's own outputs.

\hypertarget{structure-and-statistical-probability-in-structuralist-theories-of-communication}{%
\subsubsection{\texorpdfstring{Structure and statistical probability in
structuralist theories of communication
}{Structure and statistical probability in structuralist theories of communication }}\label{structure-and-statistical-probability-in-structuralist-theories-of-communication}}

These experiments with \emph{Ulysses} and prompt engineering reveal an
interplay between probabilities and structure that has a long history,
and was a specific topic of interest in the postwar convergence of
theories of language and computation. In the history of linguistics, the
relationship between language as a complex system of rules and language
as a tool for communication received a particular importance during the
Cold War and even before that, in the work of war-time encryption,
decryption and computation of scholars like Alan Turing (Edwards, 1996).
In an intentional way, military science played a key role in furthering
research in this field and bringing together a logistical concern with
communication that delved into the physical mechanics of transferring,
receiving and interpreting signals with some of the emerging new
theories in linguistics.

An interesting case in this regard, often overlooked in English-speaking
literature, was the Moscow School established in 1956. The school
brought together some of the brightest minds of Soviet linguistics at
the time, among which Pyotr Savvich Kuznetsov, Vyacheslav Vsevolodovich
Ivanov, Isak Yosifovich Revzin, Boris Andreyevich Uspenskiy, with key
mathematicians like Viktor Aleksandrovich Uspenskiy, Roland Ljovovich
Dobroshin and Olga Kulagina (Revzin 1977). The linguists from the Moscow
School were part of the strong tradition in structural linguistics
developing in the USSR at that time that delved into research on
poetics, folklore and mythology. Structuralism, with its root in the
linguistic theory of the Swiss scholar Ferdinand de Saussure, whose
\emph{Course in General Linguistics} provided a methodological and
conceptual framework, soon became a dominant epistemological approach in
the social sciences and humanities. With the seminal works in semiotics
by Roman Jacobson, in anthropology by Claude Lévi-Strauss, in folklore
by Vladimir Propp, in psychoanalysis by Jacques Lacan, and even, to some
extent, sociology by Émile Durkheim, the predominant paradigm around the
mid-20th century was concerned with structure, binary oppositions and
deviation.

At the Moscow School, however, the pairing of structural linguists and
mathematicians introduced a specific inflection in this intellectual
inquiry, which tried to understand the relationship between structure,
statistical probability and computation. In his overview of the state of
Soviet structural linguistics, Isaak Revzin (1977) lists the key tasks
addressed by the Moscow school as: 1) the development of machines
capable of automatic analysis of natural language; 2) the development of
a compact informational logical device that can store information and
quickly retrieve it in response to queries; 3) perfecting the modes of
transfer of information via telephone, telegraph and radio channels; 4)
the development of a device that can process speech and record it in
written form (an automated typist); and 5) machine translation from one
language to another.

To a large extent these practically-oriented tasks before the Soviet
linguists informed the line of inquiry of the school and the close
collaboration between linguists, mathematicians and physicists. These
tasks presented the linguists at the School with a very different
context of studying language. On the one hand, adhering to the
Saussurean structural school of linguistics, they imported distinctions
between linguistic levels (phonetics, syntax, semantics), and the basic
structures of language or \emph{langage} itself: the contrast between
its background rules and lexicon (\emph{langue}), and its specific
instances or utterances (\emph{parole}). On the other hand, Moscow
School linguistics were influenced by the approach of their colleagues
in mathematics and physics (notably, Andrey Markov, whose work on the
probabilities in transitions between states in algorithmic processes
(1906) dates already from the first decade of the 20th century).
Increasingly, the interest in Soviet physics and mathematics was
focusing on the rules of prediction in the construction of utterances
and communication, which, in the context of the technological
innovations catalysed by World War II and accelerated by the militarised
competition of the Cold War, translated to research on the efficient
transmission and decoding of communication. Such was, for example, the
focus of research of Mark Dolukhanov, who applied statistics to the
study of technologically-mediated communication (1955). His concern with
the efficiency of technological transmission of human communication
formulated the frequency of occurrences of individual phonemes as a
solution to the problem of loss, interference and noise in transmission
channels. If knowledge of statistical probability could predict the
likelihood of occurrence and co-occurrence of phonemes, there could be a
technological mode of disambiguation, which would prevent
misunderstanding and what Claude Shannon (1948) referred to as `entropy'
of information. Dolukhanov, however, insisted on the role of internal
language structure and rules for predicting the likelihood of occurrence
of certain combinations of letters and phonemes. He saw an
interdependence between internal structure and surface statistics, where
the structure limits the possibility of occurrence of certain
combinations but where this structure is, in turn, deduced through
statistical analysis of the language.

This new approach to language led to experiments with the significance
of statistical probabilities and co-occurrence in language, for
instance, in the work of the linguist Lev Rafailovich Zinder, who
incorporated the notion of probability in his own research on language.
Zinder (1958) argued that in linguistic utterances some sequences of
phonemes and lexemes are more probable than others. The degree of
probability is determined by grammatical and lexical (or semantic)
characteristics. An example of grammatically determined degree of
probability of co-occurrence would, for instance, be the coordination
between words in terms of gender, case and singular/plural in the
Russian language. Semantically determined probability, however, is much
more challenging to establish. It is defined by the meaning of words and
their likelihood of being used together -- for example, `red flag',
`sunny day' or `starry night'. Zinder's own work focused on the
probability of co-occurrence of sequences of phonemes in Russian, a task
that was well-placed within the imperatives for efficient communication
via technological channels pursued in the context of the Cold War. This
echoes and no doubt responds to the more famous work on information
theory by Turing, Shannon, and others, but places much greater influence
on the continued relevance of Saussurean structure.

Ironically, Zinder's programme for understanding semantic probability of
co-occurrence, inspired by Dolukhanov's physical theory of information,
in turn found a different interpretation in the USA, where, around the
same time, a group of psychologists, Charles Osgood, George Suci and
Percy Tannenbaum (1957) were trying to develop a mathematically-backed
measure of meaning in language through quantifying the likelihood of
semantic co-occurrences. Their development of multidimensional measures
of semantic proximity and likelihood of co-occurrence was later advanced
in the work of preparing the semantic network WordNet (Miller et al
1990), and in the use of semantic embeddings and vectorisation that
today forms the basis for the efficiency of LLMs like ChatGPT.

The endeavour to quantify language and develop a statistical theory of
communication and linguistic usage, however, was concerned (and is
concerned), in much broader sense, with the relationship between
structure and statistical probability. Despite the advances in
computation, WordNet and ChatGPT represent only single sides of this
relationship -- the first modelling semantic structure, the second token
likelihood (from which an emerging latent structure may still diverge
from human expectations, and so require alignment). Albeit from a
different perspective, scholars like Dolukhanov, Zinder and Osgood
interrogated the possibility of existence of an underlying, implicit
structure in language that they assumed to be part of the cognitive
schemata that enables humans to produce and understand linguistic
utterances. Dolukhanov (1955) uses the notion of implicit understanding
on the side of the receiver which is weaved into a theory of
communication concerned with efficiency and entropy, or the loss of
information. His use of the notion of implicit rules (structure) comes
in connection with this concern. The implicit rules on the side of the
receiver of information make it easier to predict and understand an
utterance and, therefore, improve the economy of communication. His
theory of information was adopted by Zinder in his work on linguistic
probabilities who espouses the theory of a dual process of communication
influenced by contemporaneous research into the transfer of
communication via electronic and telegraphic/telephone channels and
issues of coding/decoding (encryption/decryption). According to this
dual theory of communication, the process of transferring information
consists of two aspects: perception (or reception), which comprises the
physical transfer of signals, and understanding, which refers to the
encoding and decoding of signals.

Thus the reconciliation of structure and statistics not only formed an
important part of the early history of experiments with mathematical
linguistics and quantification of linguistic use, which precede and
inform the development of current LLMs, but it was also inextricably
woven into a linguistic theory of mind. Osgood and colleagues, for
example, were heavily influenced in their research on semantic
quantification by an implicit assumption of underlying mental frames of
reference in the production of meaning. Their method of semantic surveys
and questionnaires relied on the use of binary oppositions (a typically
structuralist model of cognitive frameworks) and the participants' own
intuitive judgement of proper language use. Zinder later collaborated
with another Soviet linguist, Nikolai Andreev, to advance an amendment
to the Saussurean conception of language precisely through the
supplementation of \emph{langue}/\emph{parole}/\emph{langage} with the
Markovian idea of `speech probability' (Andreev \& Zinder 1964).
`Probability' here operates across a `hierarchically organised and
multi-dimensional structure', and also helps to account for
individualised variation of speech acts, such as authorial style and
occupational dialects, within the constraints of a wider inherited
language.

In light of contemporary debates about the status of
statistically-informed language models, which often juxtapose the
production of utterances through statistical probability to true meaning
and comprehension -- famously through the critique of LLMs as stochastic
parrots (Bender et al 2021) -- our revisiting of the history of
quantification of semantics suggests a more complicated picture. In the
works of Dolukhanov, Zinder, Osgood and others, implicit language
structures and surface statistical variations are seen in relation to
each other: the hope of researchers is that by studying the latter they
can gather some insight about the mechanisms and rules of -- as well as
deviations from --~the former. This interdependency is also grounded in
the need to reconcile a linguistic theory of the mind with the
communicative aspect of language, i.e. the mechanisms through which
implicit language rules and structures are intelligible and shared among
language users.

A speculative question that emerges from this work in relation to LLMs
is whether and to what degree they learn an implicit structure of
language via purely statistical methods. WordNet, a Princeton University
project that represents lexical entries in terms of their semantic
relations of synonymity, antonymy and generality (hypernymy) to other
entries, exemplifies the contrastive structure that follows from a
broadly Saussurean program. LLMs, on the other hand, learn an implicit
or latent structure, one that maps and compresses the examples of
language use they are trained upon. The dimensions of this structure may
or may not correspond to human intuition (e.g. distinctions between
nouns and verbs, concrete and abstract nouns, and so on), and may
arbitrarily confuse or interleave syntactic with semantic distinctions.
In one sense such models realise the ambitions of the Moscow School
--~latent structure does emerge through recurrent attention to textual
tokens and their relations, and that structure appears to exhibit
hierarchical relations (e.g. between syntactic and semantic levels). In
another, they frustrate those ambitions, since this emergent structure
is only what enables a model to best approximate its goal of language
generation, and may have limited relation to the structures that
underpin human communication. Since the latent machinic structure is
itself only a collection of unlabelled vectors --~sequences of numbers
--~even identifying the meaning of its dimensions requires human
analysis and interpretation. Yet the extent of parallelism between
theories of semantic structure and the latent structures learned by
machines do suggest possibilities for greater alignment over time, as
computational architectures evolve.

Whether acknowledging this relationship tilts the scales in the
direction of the possibility of autonomous linguistic production and
some degree of consciousness in LLMs is out of the scope of our current
inquiry. An important strain of current LLM research (see LeCun, 2022)
argues for the need for modules or components with different organising
principles, including an innate or at least differently-acquired set of
dominant semantic concepts, to supplement the unsupervised acquisition
of semantic relations from large training sets. More relevant to our
study here, the inquiries exploring the relationship between structure
and statistical variations in language in the history of computational
linguistics and LLM research also inform a particular context of the
notion and practice of alignment. First, they describe what implicit
rules are encoded in the way LLMs operate with language and how these
rules create a specific situation of linguistic coercion through the
interface of the chat function and the logic of the prompt. Second, they
also point to a fundamental difficulty encountered by the collapse of
all structure into the pure regulating influence of probabilities
--~without structure, the solution of alignment problems has no recourse
but to the level at which those probabilities operate. The
hyperdimensionality of LLMs is in this sense a misnomer: from the point
of view of alignment, all of these dimensions orient from a single point
of origin, and a single standpoint from which decisions need to be made.

\hypertarget{linguistic-structure-and-theory-of-the-mind}{%
\subsubsection{\texorpdfstring{Linguistic structure and theory of the
mind
}{Linguistic structure and theory of the mind }}\label{linguistic-structure-and-theory-of-the-mind}}

The relationship between structure and statistics reverberates in a
different way in later accounts of language and its relation to
computation. Though not directly related to the Moscow School, the
arguments of Searle, Chomsky and Kristeva qualify the relation of
language to mind, and pose enduring challenges to the alignment of a
technology to psychosocial structures that, in these arguments, remain
irreconcilable to it.

In 1980 John Searle (1980) proposed a thought experiment. A person with
no knowledge of Mandarin is locked in a room. Through the door, a native
Mandarin speaker passes notes to the person inside. Equipped with a set
of rules written in English that instruct him how to respond to each
phrase, the person inside the room successfully communicates with the
Mandarin speaker. Searle postulates that to anyone outside of the room,
it would appear that the person inside does, in fact, understand
Mandarin but this is not the case. Just like computers, even when they
perform tasks that give the appearance of consciousness and
intelligence, they are simply following a set of rules. Searle insisted
that true understanding, consciousness and intelligence required more
than the ability to respond adequately to a given context.

The Chinese room problem remains a model of thinking about the
possibility of machinic consciousness and intelligence, to the extent
that it has been recently replicated in an experiment where an AI system
designed for playing board games, Othelo.AI, has access to data about a
series of movements on the board performed by players but has no
preexisting knowledge of the board outline or the rules of the game. It
collects data about the movements and tries to predict the next move of
a player in the game. The question that researchers ask in this
experiment is whether Othello.AI constructs a world model, which in this
case would be a model of the playing board, in order to predict possible
moves and how they will affect the course of the game (Li et al 2023).

These debates around the status of linguistic expression in LLMs are not
confined strictly within the domain of computational science but have
already prompted discussions about the impact of such
statistically-based models on our theoretical understanding of language.
Steven Piantadosi (2023) claims that ChatGPT and similar models are
refuting core theories in the field of linguistics that underpin the
relationship between human thought and linguistic expression. He focuses
in particular on Noam Chomsky's key theoretical argument that grammar is
independent of meaning and `probabilistic models give no particular
insight into some of the basic problems of syntactic structure' (Chomsky
1957, p. 17). Chomsky insists that grammar exists as an underlying
complex system of rules and relations that is abstract and separate from
discrete lexical expressions. In a sense, Chomsky, like Searle, insists
on a pre-lexical phase of language that exists outside of the
perceptible plane of expression, a notion that presupposes a separation
of the communicative function of language from a function that is
entirely dedicated to abstract reasoning and cognition. The implication
of such understanding of language is that constructing intelligible
sentences and successfully conducting a dialogue is not enough proof of
high-level cognition.

Among the scholars adopting a view of language as a dual system of
communication and abstract reasoning is the feminist literary theorist
and psychoanalyst Julia Kristeva. In her work on the semiotics of
language (Kristeva 1980), she juxtaposes symbolic to semiotic language,
the latter of which is seen as embodied relations, affects and desires
that an infant allegedly naturally acquires through their mother.
Kristeva's position is distinguished from Chomsky's and Searle's, in
that the speaking subject is always and necessarily split between these
semiotic -- corresponding to drives and their `orderings' -- and
symbolic -- corresponding to rule-following communication --~registers.
Drawing from Freudian-Lacanian psychoanalysis, in Kristeva's conception
there is no `metaphysical foundation', no `consciousness as a
synthesising unity' (Kristeva 2002, p. 60). Yet it is remarkable to what
extent Kristeva, Chomsky and Searle, despite their theoretical and
disciplinary differences, coming from different schools of thought all,
in their own way, assume the existence and relevance of an underlying
non-communicative and, to some extent, non-linguistic, function of
language, one that cannot be comprehended or discerned from its mere
usage but is innate, intuitive and hard to replicate. Chomsky's later
work on language continues to insist on the existence of two separate
planes of language. In it, he distinguishes between language faculty in
the narrow sense, which he refers to as `the abstract linguistic
computational system', independent of the `sensory-motor' and
`conceptual-intentional' systems that, combined, constitute language
faculty in the broad sense (Hauser, Chomsky \& Fitch 2002).

There are at least two ways in which these linguistic arguments can be
counterposed to large language models. The first is, as discussed above,
the absence of structure in these models. Tokens in a model are related
to other tokens, and in addition are marked (in Transformer-based models
like GPT) positionally -- at what position they occur in a sentence or
equivalent syntactic structure. Otherwise these tokens are unmarked by
grammatical categories: a language model has no record of whether tokens
are objectively nouns, verbs, and so on. This differs entirely from
Chomsky's account of how language forms in human cognition, via an
arrangement of grammatical parts into noun phrases, verb phrases, and so
on. In a language model, syntactical correctness is instead purely a
function of inductive regularisation during training: in English
language sentence completions, a verb will follow a subject noun just
because it has done so most of the time in the preceding million
sentences it has encountered.

This probabilistic approach to sense-making has provoked the
neo-Searlian or neo-Chomskian scepticism voiced by AI critics (including
Bender et al. (2023); see also Munn et al. (2023)). GPT-4 can
more-or-less `pass' Chinese-room style experiments because of the scale
of its training, the volume of human feedback and the ingenuity of its
architects. There is uncorroborated evidence for instance that GPT-4
employs a `mixture-of-experts' architecture, which embeds an
approximation of an `inner voice' that critiques and filters model
predictions, improving accuracy without reference to an externalised
ground truth (Liu et al. 2023). However even this architecture assumes
simply more neural networks -- a sequence of GPTs checking each other's
outputs --~without the kind of hierarchical differentiation that
separates, in Searle's view, deep semantic understanding from symbolic
manipulation, or in Chomsky's view, linguistic competence on the basis
of grammatical categories from imitative performance.

A Kristevan critique however might operate on a different basis.
Alongside rule following and proposition testing, language production
takes place in a subject embodied and related to a set of spatio-social
conditions. A baby's act of crying for its mother, and for the breast
that represents reassurance and nourishment, is not a circumstantial
step towards language acquisition, but a response to a biological and
psychological drive that remains present even as this nascent subject
develops more supple linguistic faculties. Here structures must be
thought beyond those embedded in language itself, and include relations
to parental beings, home, food, pleasure, and as the subject develops,
its own sense of its temporal horizons and selfhood. Language is caught
up within, while also serving to condition the subject's desire.
Subjects without bodies and biographies are not properly subjects at
all, but shallow incantations of symbols (Magee et al 2023).

\hypertarget{conclusion}{%
\subsubsection{Conclusion}\label{conclusion}}

In relation to the problem of alignment, we argue the work of the Moscow
school constituted an early effort to devise a system of communication
that integrated deep structure and shallow statistics. The later
arguments of Searle, Chomsky and Kristeva, despite strong differences,
affirm the presence of a structure that extends beyond language into
psychic and social conditions of human experience. The success and
limits of recent LLM research has resurfaced this general challenge in a
modified form. LeCun's whitepaper (2022) for example suggests that
research into early childhood psychology can assist the design of
`autonomous machine intelligence', addressing the limits of LLMs to
develop common sense developed from `direct experience with an
underlying reality'. Schmidhuber (2023; 1990) suggests in response that
his own work from the 1980s had already advanced ideas of a `world
model' that checks and constrains predictions of a separate `controller'
component. Structure here is however limited to the arrangement of
computing components; nothing `structural' about the world, the
communicative situation or an embodied mind is pre-given to these
components. Neuro-symbolic systems (e.g. Sarker et al. 2021) couple
artificial neural networks with a symbolic system that reasons over
`expert knowledge', in the form of databases or ontologies,~to ground
predictions. Such hybrid systems include the equivalent of the `rules'
that Dolukhanov suggested were encoded in the receiver of information,
and the separate functions of the neural and symbolic subsystems
correspond, despite the shift in nomenclature, to what Zinder identified
respectively as the perceptual / receptive and understanding aspects of
communication. These neuro-symbolic systems can be seen perhaps as a
realisation then of models first proposed by the Moscow school.

Engineers and contractors at companies like OpenAI implicitly impart a
form of deep structure to language models like ChatGPT. Reinforcement
learning from human feedback (RLHF) seeks to adjust model predictions to
adhere to a priori principles of `helpfulness', `truth', and
`harmlessness' (Ouyang et al. 2022). Our experiments illustrate that
such efforts at alignment remain in the space of probabilistic
prediction that remains, despite the hyperdimensionality of language
models, one-dimensional. Alignment here is, in other words, modification
of vectors to a singular set of variables expressed as single-termed and
flattened out principles. \emph{Ulysses} is reinterpreted accordingly
through processes of abstraction and functional reduction. Here the
sense of Steyerl's critique of models as averaging devices appears --
less in the nature of the models themselves, than in the corrective
measures to normalise them to some pre-imagined human values.

Our argument is not that language models are flawed, or that alignment
efforts are misguided. Rather the characterisation of the problem of
alignment appears simplified. It is similar in this respect to early
medical efforts to correct and normalise the aberrant, deviant human
subject (Canguilhem 2012; Foucault 2003). Terms like `hallucination'
already signal how LLMs are conceived as both a subject, and a subject
that is pathological -- other technologies such as hammers, washing
machines or smartphones may break, but they do not hallucinate. This
language already signals a simple dichotomous target for model
remediation: to prevent hallucination is to prevent falsity, or to
produce truth. Similar to Canguilhem's (2012) presentation of nineteenth
century medicine's approach to illness, addressing language model
anomaly or deviancy involves only addressing quantitative intensities,
via the perturbation of vector embeddings.

Structural and pragmatist accounts of language point instead to the more
variegated nature of the alignment problem. Kristeva (1980) for example
distinguished the semiotic `orderings' of the pre-linguistic infant from
the symbolic laws that govern the child as it enters into speech. The
speaking subject thereafter always has to navigate between two systems:
their own desires, and the expectations of a society they are born into.
Speech act theory (Searle 1980) had also separated utterances in terms
of their effects: the transmission of information (locutionary force);
what is intended in that transmission (illocutionary force); and what
may sometimes be enacted through that transmission (perlocutionary
force).

The excerpts from Joyce exemplifies the kind of \emph{avant garde}
writing that Kristeva (1980) saw as enabling the semiotic \emph{chora}
to break through the regulating effects of language. In the pragmatic
register, it does (in fact many) things with words. These effects depend
upon an anticipated supple interpretative structure on the part of the
receiver / reader of these texts: able to hear the sound as well as
understand the sense, able to admire the beauty of the rhythm of prose
that depends upon offensive language for that rhythm, and so on. The
effect of aligned models to summarise or describe these texts results in
the reduction of ambivalence and a corresponding diminution of meaning.
In seeking to communicate something about these texts, the aligned model
also does not communicate, because what it is aligned to remains at the
same level as its internal representation: a set of numeric weights.

Last, we argue that the revision of historical debates and studies on
the relationship between structure and statistics in language production
and comprehension, suggests a new and productive angle to the debates
about the level of perceived comprehension in LLMs. As we have
demonstrated here, discussions about the relationship between language
and consciousness, in the works of authors as diverse from each other as
Chomsky, Kristeva, Searle and Dolukhanov, have been dominated by the
assumption of the existence of an internal structure, which can be
mental, affective or sensory. Such theories of internal language
structure assume a relationship between a monadic autonomous subject and
the production and decoding of meaning and are heavily anthropocentric
informing our understanding of the relationship between human language
and technology. However, this narrative juxtaposing human language
comprehension as an innate skill to the machinic production of texts,
obscures a long line of experiments and theoretical discussions that
interrogate the possibilities of mutual convergences between language,
technology and the communication of meaning. The legacy of the Moscow
School, in this regard, is especially revealing for the work done on
uncovering the technological and mechanistic aspects of structure and
meaning in the production, transmission and decoding of utterances.
Their adoption of statistics in the study of language was not just an
attempt to quantify speech. Rather, it highlighted the communicative
plane of language -- the medium where utterances are expressed outside
of the monadic entity of the thinking, sensing or feeling subject, and
have to navigate the mechanisms of efficiently transmitting information.
Our research traces the development of LLMs exactly to this tradition of
researching and experimenting with language in its communicative
function. An important consequence of this legacy is the logic of
alignment that we observe through our experiments.

The attempts at alignment are framed through this communicative function
of language, where language escapes the domain of individual
subjectivity and enters the domain of operationality. We see this
context of analysing LLMs as especially poignant, as it shifts the focus
from consciousness to efficiency of transmission and from subjectivity
to intersubjective relations. It is exactly in these attempts to align
texts produced by the models and text produced by users that we see the
most disruptive consequences of LLMs. Rather than evaluating their level
of consciousness or human-like comprehension, we argue that it is their
role in steering language use in a particular direction, through the
imperative of mutual intelligibility and economy of communication, that
has the most profound effects on the relationship between language, mind
and technology.

\hypertarget{conflict-of-interest-statement}{%
\paragraph{\texorpdfstring{\textbf{Conflict of Interest
Statement}}{Conflict of Interest Statement}}\label{conflict-of-interest-statement}}

On behalf of all authors, the corresponding author states that there is
no conflict of interest.

\textbf{Data Availability Statement}

All data used in this article is `synthetic', generated by AI systems
such as ChatGPT. Due to the stochastic nature of LLM and other AI
interactions, it is not possible to replicate the exchanges and outputs
of our experiments precisely.

\hypertarget{references}{%
\subsubsection{References}\label{references}}

Andreev ND, Zinder LR (1964) On the notions of the speech act, speech,
speech probability, and language. Linguist 2(4):5-13.
\url{https://doi.org/10.1515/ling.1964.2.4.5}

Bender EM, Gebru T, McMillan-Major A, Shmitchell S (2021) On the dangers
of stochastic parrots: Can language models be too big?. In:
Proceedings of the 2021 ACM conference on fairness, accountability, and
transparency, pp 610-623.
\url{https://doi.org/10.1145/3442188.3445922}

Beurer-Kellner L, Fischer M, Vechev M (2023) Prompting is programming: A
query language for large language models. Proc ACM Program Lang
7(PLDI):946-1969.
\url{https://doi.org/10.1145/3591300}

Canguilhem G (2012) On the normal and the pathological (Vol 3). Berlin,
Springer Science \& Business Media

Castelvecchi D (2016) Can we open the black box of AI? Nature
538(7623):20.
\url{https://doi.org/10.1038/538020a}

Chalmers DJ (2023) Could a large language model be conscious?
\url{https://doi.org/10.48550/arXiv.2303.07103}

Chang Y, Wang X, Wang J, Wu Y, Yang L, Zhu K, Chen H, Yi X, Wang C, Wang
Y, Wei Y, Zhang Y, Chang Y, Yu PS, Yang Q, Xie X (2023) A survey on
evaluation of large language models.
\url{https://doi.org/10.48550/arXiv.2307.03109}

Chomsky N (1957) Syntactic Structures. Mouton de Gruyter.

Chun WHK (2005) On software, or the persistence of visual knowledge.
Grey Room 18:26-51.
\url{https://doi.org/10.1162/1526381043320741}

Chun WHK (2016) Updating to remain the same: Habitual new media.
Cambridge MA, MIT Press

Coulter D (2023) AI experts disown Musk-backed campaign citing their
research. Reuters.
\url{https://www.reuters.com/technology/ai-experts-disown-musk-backed-campaign-citing-their-research-2023-03-31/}.
Accessed 28 December 2023

DAIR.AI (2023) Examples of prompts.
\url{https://www.promptingguide.ai/introduction/examples}.
Accessed 28 December 2023

Dolukhanov M P (1955) Vvedenie v teoriju peredachi informacii po
elektricheskim kanalam svjazi. Svjazisdat.

Durkheim E (2005) Suicide: A study in sociology. London, Routledge

Edwards PN (1996). The closed world: Computers and the politics of
discourse in Cold War America. Cambridge MA, MIT Press

Foucault M (2003) Madness and civilization. London, Routledge

Geoghegan B D (2022) \emph{Code: From Information Theory to French
Theory}. Duke University Press.

Google (2023) Google search help.
\url{https://support.google.com/websearch/answer/2466433?hl=en}.
Accessed 2 December 2023

Harwell D (2023) Tech's hottest new job: AI whisperer. No coding
required. Washington Post.
\url{https://www.washingtonpost.com/technology/2023/02/25/prompt-engineers-techs-next-big-job/}.
Accessed 28 December 2023

Hauser MD, Chomsky N, Fitch WT (2002) The faculty of language: what is
it, who has it, and how did it evolve? Sci 298(5598):1569-1579.

Jakobson R (1956) Two aspects of language and two types of aphasic
disturbances. In: Jakobson R, Halle M Fundamentals of language (vol 1),
Berlin, Walter de Gruyter, pp. 69-96.

Kristeva J (1980) Desire in language: A semiotic approach to literature
and art. New York, Columbia University Press

Kristeva J (2002) The portable Kristeva. New York, Columbia University
Press

LeCun Y (2022). A path towards autonomous machine intelligence. Open
Review, 62.

Li K, Hopkins AK, Bau D, Viégas F, Pfister H, Wattenberg M (2022)
Emergent world representations: Exploring a sequence model trained on a
synthetic task.
\url{https://doi.org/10.48550/arXiv.2210.13382}

Liu B, Ding L, Shen L, Peng K, Cao Y, Cheng D, Tao D (2023) Diversifying
the mixture-of-experts representation for language models with
orthogonal optimizer.
\url{https://doi.org/10.48550/arXiv.2310.09762}

Magee L, Arora V, Munn L (2023) Structured like a language model:
Analysing AI as an automated subject. Big Data \& Society, 10(2)
\url{https://doi.org/10.1177/20539517231210273}

Markov A A (1906) Rasprostranenie zakona boljshikh chisel na velichiny,
zavisjashtie drug ot druga. Izvestija Fiziko-matematicheskogo obshtestva
pri Kazanskom universitete, pp.135-156.

Miller GA, Beckwith R, Fellbaum C, Gross D, Miller KJ (1990)
Introduction to WordNet: An on-line lexical database. Int J Lexicogr
3(4):235-244.

Munn L, Magee L, Arora V (2023) Truth machines: Synthesizing veracity in
AI language models. AI \& Soc.
\url{https://doi.org/10.1007/s00146-023-01756-4}

Osgood, C E, Suci, G J, \& Tannenbaum, P H (1957) \emph{The measurement
of meaning} (No. 47). University of Illinois press.

Ouyang L, Wu J, Jiang X, Almeida D, Wainwright C, Mishkin P, Zhang C,
Agarwal S, Slama K, Ray A, Schulman J (2022) Training language models to
follow instructions with human feedback. Adv Neural Inform Process Syst
35:27730-27744.
\url{https://doi.org/10.48550/arXiv.2203.02155}

Roose K (2023) Bing's AI chat: `I Want to Be Alive'. New York Times.
\url{https://www.nytimes.com/2023/02/16/technology/bing-chatbot-transcript.html}.
Accessed 28 December 2023

Parisi L (2019) The alien subject of AI. Subjectiv 12:27-48.
\url{https://doi.org/10.1057/s41286-018-00064-3}

Patel D, Raut G, Zimlichman E, Cheetirala S, Nadkarni G, Glicksberg BS,
Freeman R, Timsina P, Klang E (2023) The limits of prompt engineering in
medical problem-Solving: A Comparative Analysis with ChatGPT on
calculation based USMLE Medical Questions. medRxiv, pp.2023-08.

Pasquale F (2015) The black box society: the secret algorithms that
control money and information. Cambridge MA, Harvard University Press

Perrigo B (2023) Exclusive: OpenAI used Kenyan workers on less than \$2
per hour to make ChatGPT less toxic. TIME.
\url{https://time.com/6247678/openai-chatgpt-kenya-workers/}
Accessed 28 December 2023

Piantadosi S (2023). Modern language models refute Chomsky's approach to
language. \emph{Lingbuzz Preprint, lingbuzz}, \emph{71-80}.

Revzin I I (1977) Sovremennaja strukturnaja lingvistika. Problemy i
metody. Moscow: Nauka.

Saba WS (2023) Stochastic LLMs do not understand language: towards
symbolic, explainable and ontologically based LLMs. In International
Conference on Conceptual Modeling (pp. 3-19). Cham: Springer Nature
Switzerland.

Sarker MK, Zhou L, Eberhart A, Hitzler P (2021) Neuro-symbolic
artificial intelligence. AI Commun 34(3):197-209.

de Saussure F (2011) Course in general linguistics. New York City,
Columbia University Press

Searle J (1980) Minds, brains and programs. Behav Brain Sci 3(3):417-24.
\url{https://doi.org/10.1017/S0140525X00005756}

Schmidhuber J (1990) Making the world differentiable: On using
self-supervised fully recurrent neural networks for dynamic
reinforcement learning and planning in non-stationary environments.
Inst. für Informatik, 126

Shannon CE (1948) A mathematical theory of communication. The Bell
System Technical Journal 27(3):379-423.
\url{https://doi.org/10.1002/j.1538-7305.1948.tb01338.x}

Steyerl H (2023) Mean images. New Left Review. 140/141, March/April
2023:82-97.
\url{https://newleftreview.org/issues/ii140/articles/hito-steyerl-mean-images}

Stiegler B (2018) Automatic society, volume 1: the future of work. New
Jersey, John Wiley \& Sons

SynapticLabs (2023) Prompt engineering: Quicksilver OS syntax.
\url{https://blog.synapticlabs.ai/quicksilver-os-syntax}
Accessed 28 December 2023

Taylor A (2018) The automation charade. Logic Magazine, 5(1)

Vaswani A, Shazeer N, Parmar N, Uszkoreit J, Jones L, Gomez AN, Kaiser
Ł, Polosukhin I (2017) Attention is all you need. Advances in neural
information processing systems, 30.
\url{https://doi.org/10.48550/arXiv.1706.03762}

Zinder L R (1958) O lingvisticheskoj verojatnosti. Voprosy jazykoznanija
VII, March-April, 121-125

\end{document}